\documentclass[10pt,twocolumn,letterpaper]{article}

\usepackage{iccv}
\usepackage{times}
\usepackage{epsfig}
\usepackage{graphicx}
\usepackage{amsmath}
\usepackage{amssymb}
\usepackage{multirow}
\usepackage{subfigure}
\usepackage{microtype}
\usepackage{booktabs}
\usepackage{color}
\usepackage{xcolor}
\usepackage{colortbl}
\usepackage{enumitem}
\usepackage{multicol}
\newcommand{\tabincell}[2]{\begin{tabular}{@{}#1@{}}#2\end{tabular}}
\newcommand{\R}{\mathbb{R}}

\usepackage{url}
\usepackage{makecell}
\newcommand*\samethanks[1][\value{footnote}]{\footnotemark[#1]}
\usepackage{amssymb}
\usepackage{pifont}
\usepackage{dsfont}
\newcommand{\cmark}{\ding{51}}%
\newcommand{\xmark}{\ding{55}}%
\usepackage[accsupp]{axessibility}  

\usepackage[breaklinks=true,bookmarks=false]{hyperref}

\iccvfinalcopy 

\ificcvfinal\fi
\definecolor{mygray}{gray}{.92}
\begin{document}

\title{Co-Scale Conv-Attentional Image Transformers}

\author{Weijian Xu\thanks{~indicates equal contribution.} \quad Yifan Xu\samethanks \quad Tyler Chang \quad Zhuowen Tu\\
University of California San Diego \\
{\tt\small $\{$wex041, yix081, tachang, ztu$\}$@ucsd.edu }
}

\maketitle
\ificcvfinal\thispagestyle{empty}\fi

\begin{abstract}
In this paper, we present Co-scale conv-attentional image Transformers (CoaT), a Transformer-based image classifier equipped with co-scale and conv-attentional mechanisms. First, the co-scale mechanism maintains the integrity of Transformers' encoder branches at individual scales, while allowing representations learned at different scales to effectively communicate with each other; we design a series of serial and parallel blocks to realize the co-scale mechanism. Second, we devise a conv-attentional mechanism by realizing a relative position embedding formulation in the factorized attention module with an efficient convolution-like implementation. CoaT empowers image Transformers with enriched multi-scale and contextual modeling capabilities. On ImageNet, relatively small CoaT models attain superior classification results compared with similar-sized convolutional neural networks and image/vision Transformers. The effectiveness of CoaT's backbone is also illustrated on object detection and instance segmentation, demonstrating its applicability to downstream computer vision tasks.

\let\thefootnote\relax\footnotetext{Code at \url{https://github.com/mlpc-ucsd/CoaT}.}

\end{abstract}

\section{Introduction}

\begin{figure}[t]
\begin{center}
\includegraphics[width=1.00\linewidth]{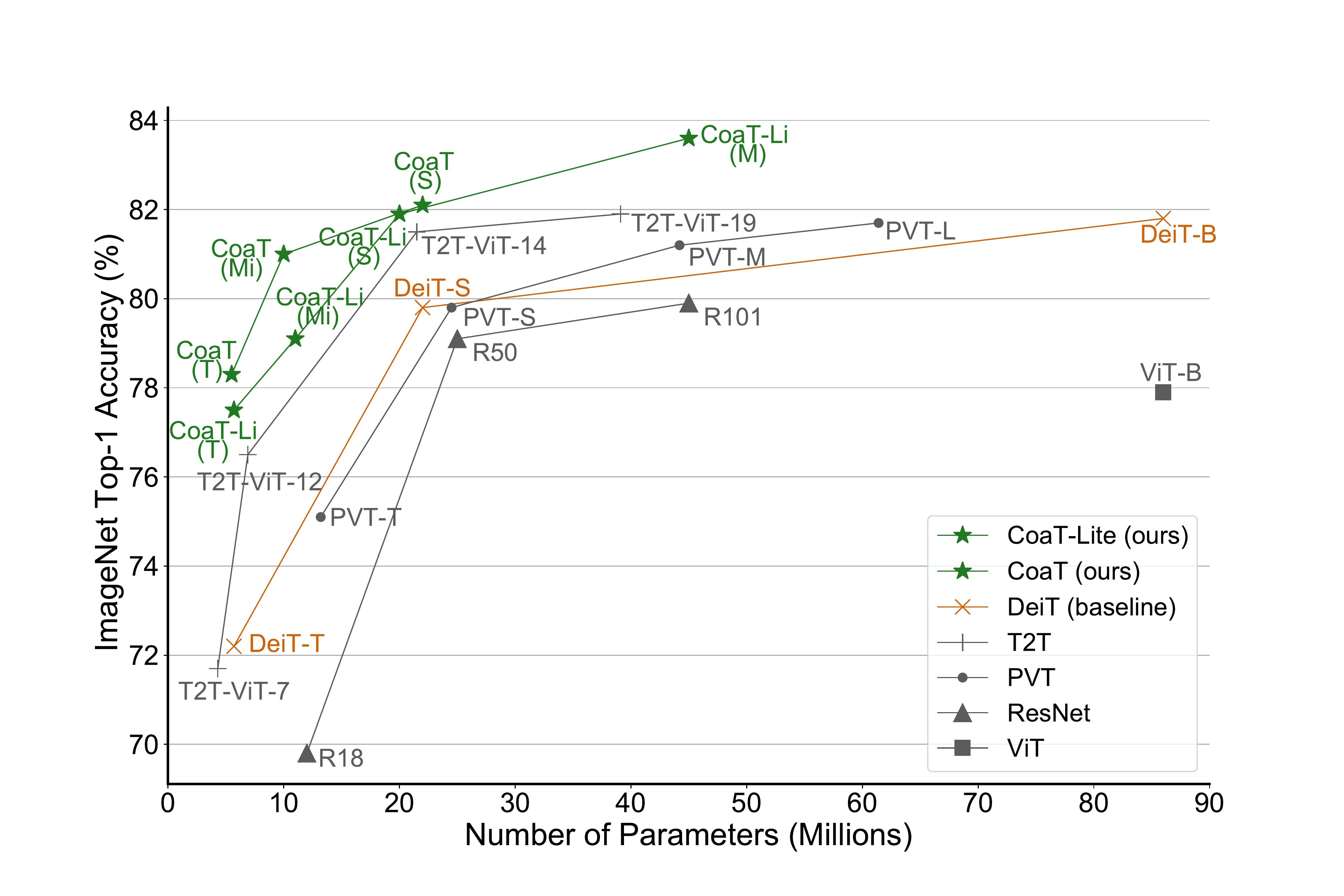}
\end{center}
\vspace{-2mm}
   \caption{\textbf{Model Size vs. ImageNet Accuracy.} Our CoaT model significantly outperforms other image Transformers. Details are in Table \ref{tab:imagenet}.}
\label{fig:model-acc}
\vspace{-3mm}
\end{figure}

A notable recent development in artificial intelligence is the creation of attention mechanisms \cite{xu2015show} and Transformers \cite{vaswani2017attention}, which have made a profound impact in a range of fields including natural language processing \cite{devlin2019bert,radford2018improving}, document analysis \cite{xu2020layoutlm}, speech recognition \cite{dong2018speech}, and computer vision \cite{dosovitskiy2020image,carion2020end}. In the past, state-of-the-art image classifiers have been built primarily on convolutional neural networks (CNNs) \cite{lecun1998gradient,krizhevsky2012imagenet,szegedy2015going,simonyan2015very,he2016deep,xie2017aggregated} that operate on layers of filtering processes. Recent developments \cite{touvron2020training,dosovitskiy2020image} however begin to show encouraging results for Transformer-based image classifiers.

In essence, both the convolution \cite{lecun1998gradient} and attention \cite{xu2015show} operations address the fundamental representation problem for structured data (e.g. images and text) by modeling the local contents, as well as the contexts. The receptive fields in CNNs are gradually expanded through a series of convolution operations. The attention mechanism \cite{xu2015show,vaswani2017attention} is, however, different from the convolution operations: (1) the receptive field at each location or token in self-attention \cite{vaswani2017attention} readily covers the entire input space since each token is ``matched'' with all tokens including itself; (2) the self-attention operation for each pair of tokens computes a dot product between the ``query'' (the token in consideration) and the ``key'' (the token being matched with) to weight the ``value'' (of the token being matched with). 

Moreover, although the convolution and the self-attention operations both perform a weighted sum, their weights are computed differently: in CNNs, the weights are learned during training but fixed during testing; in the self-attention mechanism, the weights are dynamically computed based on the similarity or affinity between every pair of tokens. As a consequence, the self-similarity operation in the self-attention mechanism provides modeling means that are potentially more adaptive and general than convolution operations. In addition, the introduction of position encodings and embeddings \cite{vaswani2017attention} provides Transformers with additional flexibility to model spatial configurations beyond fixed input structures.

Of course, the advantages of the attention mechanism are not given for free, since the self-attention operation computes an affinity/similarity that is more computationally demanding than linear filtering in convolution. The early development of Transformers has mainly focused on natural language processing tasks \cite{vaswani2017attention,devlin2019bert,radford2018improving} since text is ``shorter'' than an image, and text is easier to tokenize. In computer vision, self-attention has been adopted to provide added modeling capability for various applications \cite{wang2018non,xie2018attentional,zhao2020exploring}. With the underlying framework increasingly developed \cite{dosovitskiy2020image,touvron2020training}, Transformers start to bear fruit in computer vision \cite{carion2020end,dosovitskiy2020image} by demonstrating their enriched modeling capabilities. 

In the seminal DEtection TRansformer (DETR)~\cite{carion2020end} algorithm, Transformers are adopted to perform object detection and panoptic segmentation, but DETR still uses CNN backbones to extract the basic image features. Efforts have recently been made to build image classifiers from scratch, all based on Transformers \cite{dosovitskiy2020image,touvron2020training,wang2021pyramid}. While Transformer-based image classifiers have reported encouraging results, performance and design gaps to the well-developed CNN models still exist. For example, in \cite{dosovitskiy2020image,touvron2020training}, an input image is divided into a single grid of fixed patch size. In this paper, we develop Co-scale conv-attentional image Transformers (CoaT) by introducing two mechanisms of practical significance to Transformer-based image classifiers.
The contributions of our work are summarized as follows:
\begin{itemize}
\setlength\itemsep{0mm}
 \setlength{\itemindent}{0mm}
    \item We introduce a co-scale mechanism to image Transformers by maintaining encoder branches at separate scales while engaging attention across scales. Two types of building blocks are developed, namely a serial and a parallel block, realizing {\bf fine-to-coarse}, {\bf coarse-to-fine}, and {\bf cross-scale} image modeling.
    \item We design a conv-attention module to realize {\bf relative position embeddings} with convolutions in the {\bf factorized attention} module that achieves significantly enhanced computation efficiency when compared with vanilla self-attention layers in Transformers.
    \end{itemize}
    Our resulting Co-scale conv-attentional image Transformers (CoaT) learn effective representations under a modularized architecture.
    On the ImageNet benchmark, CoaT achieves state-of-the-art classification results when compared with the competitive convolutional neural networks (e.g. EfficientNet \cite{tan2019efficientnet}), while outperforming the competing Transformer-based image classifiers \cite{dosovitskiy2020image,touvron2020training,wang2021pyramid}, as shown in Figure \ref{fig:model-acc}.


\section{Related Works}

Our work is inspired by the recent efforts \cite{dosovitskiy2020image,touvron2020training} to realize Transformer-based image classifiers. ViT \cite{dosovitskiy2020image} demonstrates the feasibility of building Transformer-based image classifiers from scratch, but its performance on ImageNet \cite{russakovsky2015imagenet} is not achieved without including additional training data; DeiT \cite{touvron2020training} attains results comparable to convolution-based classifiers by using an effective training strategy together with model distillation, removing the data requirement in \cite{dosovitskiy2020image}. Both ViT \cite{dosovitskiy2020image} and DeiT \cite{touvron2020training} are however based on a single image grid of fixed patch size.

The development of our co-scale conv-attentional transformers (CoaT) is motivated by two observations: (1) multi-scale modeling typically brings enhanced capability to representation learning \cite{he2016deep,ronneberger2015u,wang2020deep}; (2) the intrinsic connection between relative position encoding and convolution makes it possible to carry out efficient self-attention using conv-like operations. As a consequence, the superior performance of the CoaT 
models shown in the experiments comes from two of our new designs in Transformers: (1) a co-scale mechanism that allows cross-scale interaction; (2) a conv-attention module to realize an efficient self-attention operation. Next, we highlight the differences of the two proposed modules with the standard operations and concepts.
\vspace{-2mm}
\begin{itemize}
\setlength\itemsep{0mm}
 \setlength{\itemindent}{0mm}
 \item \textbf{Co-Scale vs. Multi-Scale}. Multi-scale approaches have a long history in computer vision \cite{witkin1984scale,lowe2004distinctive}. Convolutional neural networks \cite{lecun1998gradient,krizhevsky2012imagenet,he2016deep} naturally implement a fine-to-coarse strategy. U-Net \cite{ronneberger2015u} enforces an extra  coarse-to-fine route in addition to the standard fine-to-coarse path; HRNet \cite{wang2020deep} provides a further enhanced modeling capability by keeping simultaneous fine and coarse scales throughout the convolution layers. In a parallel development \cite{wang2021pyramid} to ours, layers of different scales are in tandem for the image Transformers but \cite{wang2021pyramid} merely carries out a fine-to-coarse strategy. The co-scale mechanism proposed here differs from the existing methods in how the responses are computed and interact with each other: CoaT consists of a series of highly modularized serial and parallel blocks to enable attention with fine-to-coarse, coarse-to-fine, and cross-scale information on tokenized representations. The joint attention mechanism across different scales in our co-scale module provides enhanced modeling power beyond existing vision Transformers \cite{dosovitskiy2020image,touvron2020training,wang2021pyramid}.
 
 \item \textbf{Conv-Attention vs. Attention.} 
 Pure attention-based models \cite{ramachandran2019stand,hu2019local,zhao2020exploring,dosovitskiy2020image,touvron2020training} have been introduced to the vision domain. \cite{ramachandran2019stand,hu2019local,zhao2020exploring} replace convolutions in ResNet-like architectures with self-attention modules for better local and non-local relation modeling. In contrast, \cite{dosovitskiy2020image,touvron2020training} directly adapt the Transformer \cite{vaswani2017attention} for image recognition. Recently, there have been works \cite{bello2021lambdanetworks,chu2021we} enhancing the attention mechanism by introducing convolution. LambdaNets \cite{bello2021lambdanetworks} introduce an efficient self-attention alternative for global context modeling and employ convolutions to realize the relative position embeddings in local context modeling.  CPVT \cite{chu2021we} designs 2-D depthwise convolutions as the conditional positional encoding after self-attention. In our conv-attention, we: (1) adopt an efficient factorized attention following \cite{bello2021lambdanetworks}; (2) extend it to be a combination of depthwise convolutional relative position encoding and convolutional position encoding, related to CPVT~\cite{chu2021we}. Detailed discussion of our network design and its relation with LambdaNets \cite{bello2021lambdanetworks} and CPVT \cite{chu2021we} can be found in Section~\ref{sec:conv_att_factorized} and \ref{sec:conv_att_conv_pos}.
 
\end{itemize}

\vspace{-3mm}
\section{Revisit Scaled Dot-Product Attention}
Transformers take as input a sequence of vector representations (i.e. tokens) $\mathbf{x}_1, ..., \mathbf{x}_N$, or equivalently $X \in \R^{N \times C}$.
The self-attention mechanism in Transformers projects each $\mathbf{x}_i$ into corresponding query, key, and value vectors, using learned linear transformations $W^Q$, $W^K$, and $W^V \in \R^{C \times C}$. Thus, the projection of the whole sequence generates representations $Q, K, V \in \R^{N \times C}$. The \textit{scaled dot-product attention} from original Transformers \cite{vaswani2017attention} is formulated as :
\begin{equation}
\label{eq:original-att}
    \text{Att}(X) = \text{softmax}\Big(\frac{QK^\top}{\sqrt{C}}\Big) V
\end{equation}

In vision Transformers \cite{dosovitskiy2020image,touvron2020training}, the input sequence of vectors is formulated by the concatenation of a class token \texttt{CLS} and the flattened feature vectors $\mathbf{x}_1, ..., \mathbf{x}_{HW}$ as image tokens from the feature maps $F \in \R^{H\times W\times C}$, for a total length of $N=HW+1$. The softmax logits in Equation~\ref{eq:original-att} become not affordable for high-resolution images  (i.e. $N \gg C$) due to its $O(N^2)$ space complexity and $O(N^2 C)$ time complexity. To reduce the length of the sequence, ViT \cite{dosovitskiy2020image,touvron2020training} tokenizes the image by patches instead of pixels. However, the coarse splitting (e.g. 16$\times$16 patches) limits the ability to model details within each patch. To address this dilemma, we propose a \textit{co-scale} mechanism that provides enhanced multi-scale image representation with the help of an efficient \textit{conv-attentional} module that lowers the computation complexity for high-resolution images.

\begin{figure}
\centering
\includegraphics[width=8.0cm]{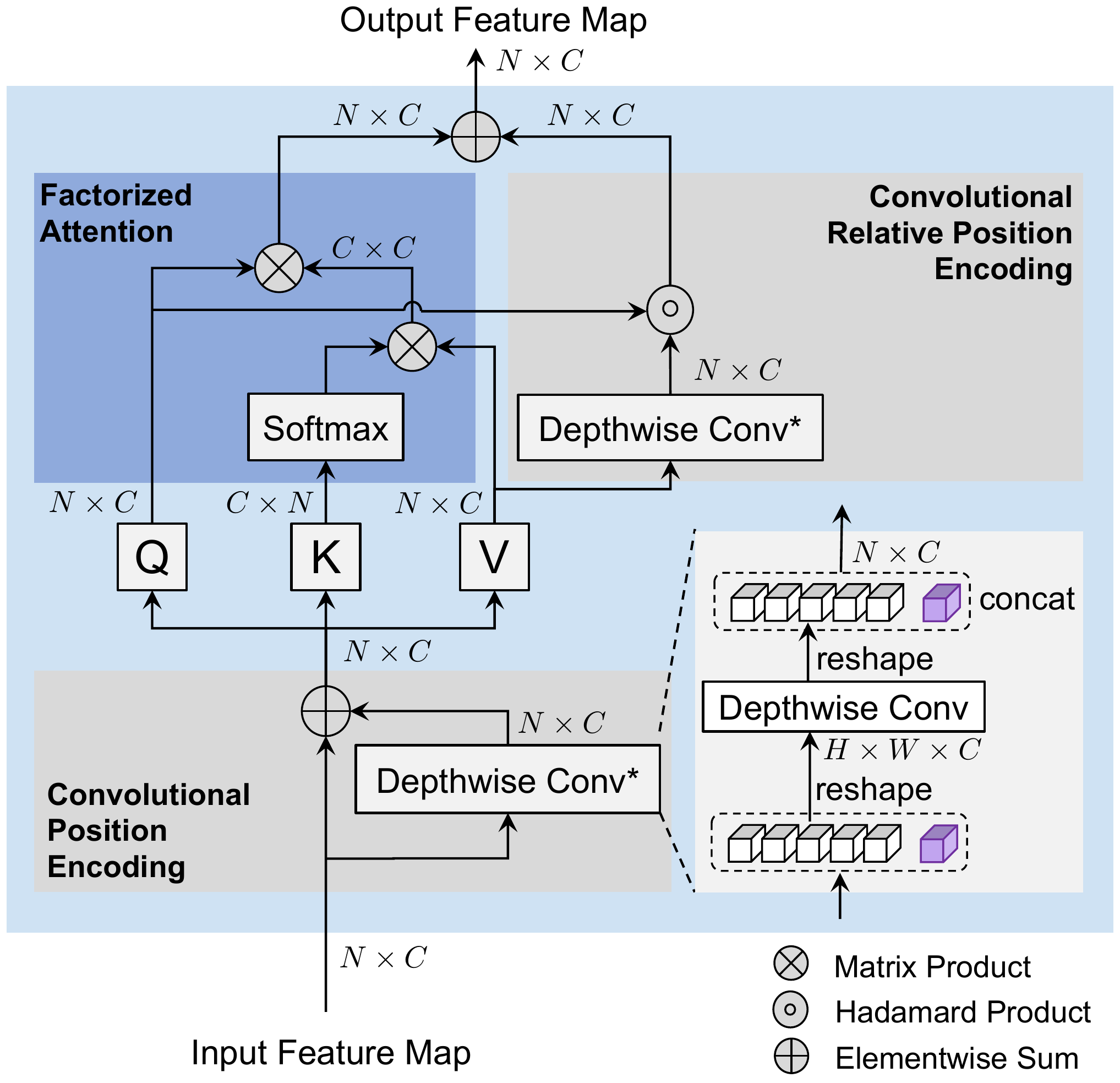}
\caption{\textbf{Illustration of the conv-attentional module.} We apply a convolutional position encoding to the image tokens from the input. The resulting features are fed into a factorized attention with a convolutional relative position encoding.}
\label{fig:conv-att}
\end{figure}

\begin{figure*}[ht]
\centering
\includegraphics[width=17.5cm]{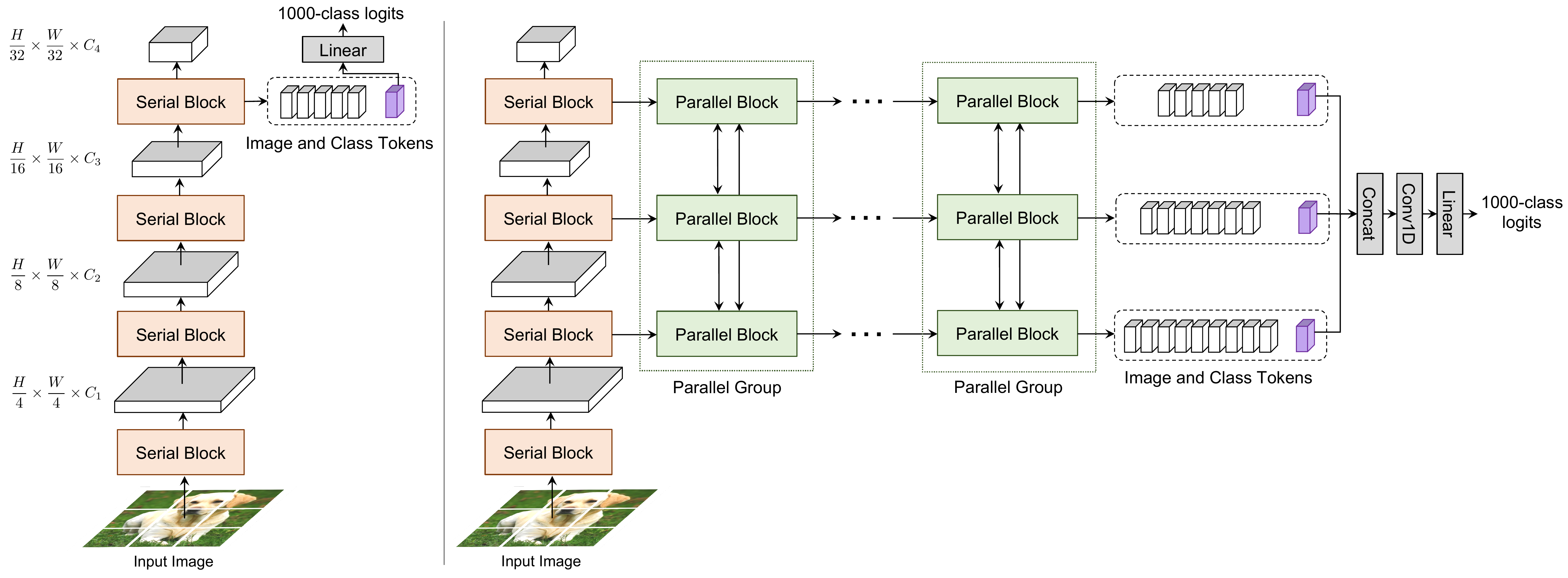}
\caption{\textbf{CoaT model architecture.} (Left) The overall network architecture of \textbf{CoaT-Lite}. CoaT-Lite consists of serial blocks only, where image features are down-sampled and processed in a sequential order. (Right) The overall network architecture of \textbf{CoaT}. CoaT consists of serial blocks and parallel blocks. Both blocks enable the co-scale mechanism. }
\label{fig:pipeline}
\end{figure*}

\section{Conv-Attention Module}
\label{sec:conv_att}
\subsection{Factorized Attention Mechanism}
\label{sec:conv_att_factorized}
In Equation~\ref{eq:original-att}, the materialization of the softmax logits and attention maps leads to the $O(N^2)$ space complexity and $O(N^2 C)$ time complexity. Inspired by recent works \cite{choromanski2020rethinking,shen2021efficient,bello2021lambdanetworks} on linearization of self-attention, we approximate the softmax attention map by factorizing it using two functions $\phi(\cdot), \psi(\cdot):\R^{N\times C}\rightarrow \R^{N \times C'}$ and compute the second matrix multiplication (keys and values) together:
\begin{equation}
\text{FactorAtt}(X) = \phi(Q)\Big(\psi(K)^\top V\Big)
\end{equation}
The factorization leads to a $O(NC'+NC+CC')$ space complexity (including output of $\phi(\cdot), \psi(\cdot)$ and intermediate steps in the matrix product) and $O(NCC')$ time complexity, where both are linear functions of the sequence length $N$. Performer \cite{choromanski2020rethinking} uses random projections in $\phi$ and $\psi$ for a provable approximation, but with the cost of relatively large $C'$. Efficient-Attention \cite{shen2021efficient} applies the softmax function for both $\phi$ and $\psi$, which is efficient but causes a significant performance drop on the vision tasks in our experiments. Here, we develop our factorized attention mechanism following LambdaNets \cite{bello2021lambdanetworks} with $\phi$ as the identity function and $\psi$ as the softmax:
\begin{equation}
\text{FactorAtt}(X) = \frac{Q}{\sqrt{C}} \Big(\text{softmax}(K)^\top V\Big)
\end{equation}
where $\text{softmax}(\cdot)$ is applied across the tokens in the sequence in an element-wise manner and the projected channels $C' = C$. In LambdaNets \cite{bello2021lambdanetworks}, the scaling factor $1/\sqrt{C}$ is implicitly included in the weight initialization, while our factorized attention applies the scaling factor explicitly. This factorized attention takes $O(NC+C^2)$ space complexity and $O(NC^2)$ time complexity. It is noteworthy that the proposed factorized attention following \cite{bello2021lambdanetworks} is not a direct approximation of the scaled dot-product attention, but it can still be regarded as a generalized attention mechanism modeling the feature interactions using query, key and value vectors.

\subsection{Convolution as Position Encoding}
\label{sec:conv_att_conv_pos}
Our factorized attention module mitigates the computational burden from the original scaled dot-product attention. However, because we compute $L = \text{softmax}(K)^\top V \in \R^{C\times C}$ first, $L$ can be seen as a global data-dependent linear transformation for every feature vector in the query map $Q$. This indicates that if we have two query vectors $\mathbf{q}_1, \mathbf{q}_2 \in \R^{C}$ from $Q$ and $\mathbf{q}_1=\mathbf{q}_2$, then their corresponding self-attention outputs will be the same:
\begin{equation}
\text{FactorAtt}(X)_1 = \frac{\mathbf{q}_1}{\sqrt{C}} L = \frac{\mathbf{q}_2}{\sqrt{C}} L = \text{FactorAtt}(X)_2
\end{equation}
Without the position encoding, the Transformer is only composed of linear layers and self-attention modules. Thus, the output of a token is dependent on the corresponding input without awareness of any difference in its locally nearby features. This property is unfavorable for vision tasks such as semantic segmentation (e.g. the same blue patches in the sky and the sea are segmented as the same category).

\vspace{-3mm}
\paragraph{Convolutional Relative Position Encoding.} To enable vision tasks, ViT and DeiT~\cite{dosovitskiy2020image,touvron2020training} insert absolute position embeddings into the input, which may have limitations in modeling relations between local tokens. Instead, following \cite{shaw2018self}, we can integrate a relative position encoding $P = \{\mathbf{p}_i\in \R^C, i=-\frac{M-1}{2},...,\frac{M-1}{2}\}$ with window size $M$ to obtain the relative attention map $EV\in \mathbb{R}^{N\times C}$; in attention formulation, if tokens are regarded as a 1-D sequence:
\vspace{-3mm}
\begin{equation}
    \text{RelFactorAtt}(X) = \frac{Q}{\sqrt{C}}\Big(\text{softmax}(K)^\top V\Big) + EV
\end{equation}
where the encoding matrix $E\in\mathbb{R}^{N\times N}$ has elements:
\begin{equation}
    E_{ij} = \mathds{1} (i,j) \mathbf{q}_i \cdot \mathbf{p}_{j-i},~~ 1\leq i,j\leq N
\end{equation}
in which $\mathds{1}(i,j) =\mathds{1}_{\{|j-i|\leq (M-1)/2\}}(i,j)$ is an indicator function. Each element $E_{ij}$ represents the relation from query $\mathbf{q}_i$ to the value $\mathbf{v}_j$ within window $M$, and $(EV)_i$ aggregates all related value vectors with respect to query $\mathbf{q}_i$. Unfortunately, the $EV$ term still requires $O(N^2)$ space complexity and $O(N^2C)$ time complexity. In CoaT, we propose to simplify the $EV$ term to $\hat{EV}$ by considering each channel in the query, position encoding and value vectors as \textit{internal heads}. Thus, for each internal head $l$, we have:
\begin{equation}
    E^{(l)}_{ij} = \mathds{1} (i,j) q^{(l)}_i p^{(l)}_{j-i},~~ \hat{EV}^{(l)}_{i}= \textstyle \sum_j E^{(l)}_{ij} v^{(l)}_j
\end{equation}
In practice, we can use a 1-D depthwise convolution to compute $\hat{EV}$:
\begin{align}
\hat{EV}^{(l)} & =Q^{(l)} \circ \text{Conv1D}(P^{(l)},V^{(l)}), \\
\hat{EV} & =Q \circ \text{DepthwiseConv1D}(P,V)
\end{align}
where $\circ$ is the Hadamard product. It is noteworthy that in vision Transformers, we have two types of tokens, the class (\texttt{CLS}) token and image tokens. Thus, we use a 2-D depthwise convolution (with window size $M\times M$ and kernel weights $P$) and apply it only to the reshaped image tokens (i.e. $Q^\text{img}, V^\text{img} \in \mathbb{R}^{H\times W\times C}$ from $Q, V$ respectively):
\begin{align}
\hat{EV}^{\text{img}} & = Q^{\text{img}} \circ \text{DepthwiseConv2D}(P,V^{\text{img}}) \\
\hat{EV} &= \text{concat}(\hat{EV}^{\text{img}}, \mathbf{0}) \\
\text{ConvAtt}(X) &= \frac{Q}{\sqrt{C}}\Big(\text{softmax}(K)^\top V\Big) + \hat{EV}
\end{align}
Based on our derivation, the depthwise convolution can be seen as a special case of relative position encoding. 

\noindent{\em Convolutional Relative Position Encoding vs Other Relative Position Encodings.} The commonly referenced relative position encoding \cite{shaw2018self} works in standard scaled dot-product attention settings since the encoding matrix $E$ is combined with the softmax logits in the attention maps, which are not materialized in our factorized attention. Related to our work, the main results of the original LambdaNets \cite{bello2021lambdanetworks} use a 3-D convolution to compute $EV$ directly and reduce the channels of queries and keys to $C_K$ where $C_K<C$, but it costs $O(NCC_K)$ space complexity and $O(NCC_KM^2)$ time complexity, which leads to relatively heavy computation when channel sizes $C_K, C$ are large. A recent update in LambdaNets \cite{bello2021lambdanetworks} provides an efficient variant with depth-wise convolution under resource constrained scenarios. Our factorized attention computes $\hat{EV}$ with only $O(NC)$ space complexity and $O(NCM^2)$ time complexity, aiming to achieve better efficiency.

\begin{figure}
\centering
\includegraphics[width=7.0cm]{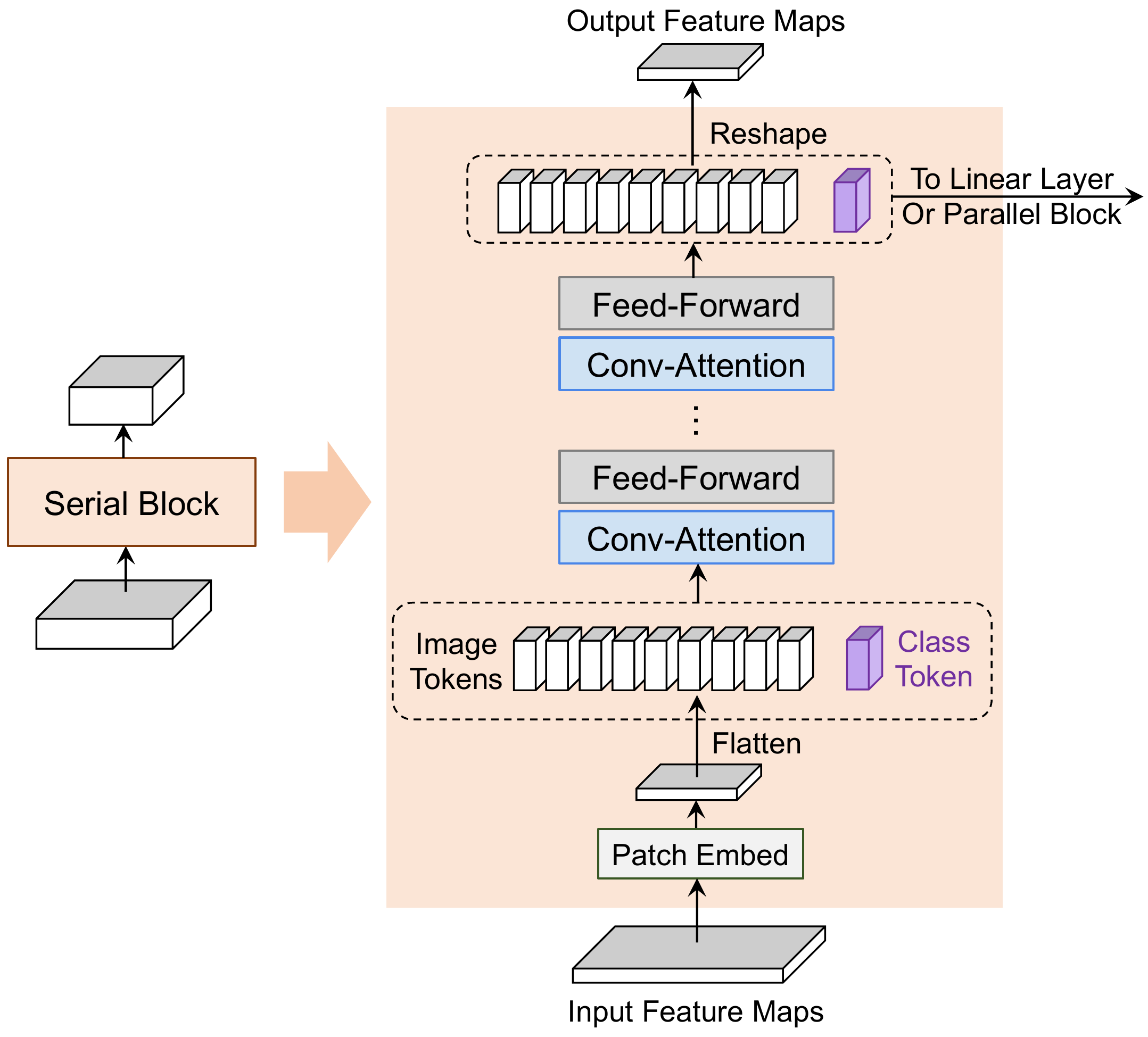}
\caption{\textbf{Schematic illustration of the serial block in CoaT.} Input feature maps are first down-sampled by a patch embedding layer, and then tokenized features (along with a class token) are processed by multiple conv-attention and feed-forward layers.}
\label{fig:serial-block}
\end{figure}

\begin{figure*}[htb]
\centering
\vspace{-3mm}
\includegraphics[width=13.5cm]{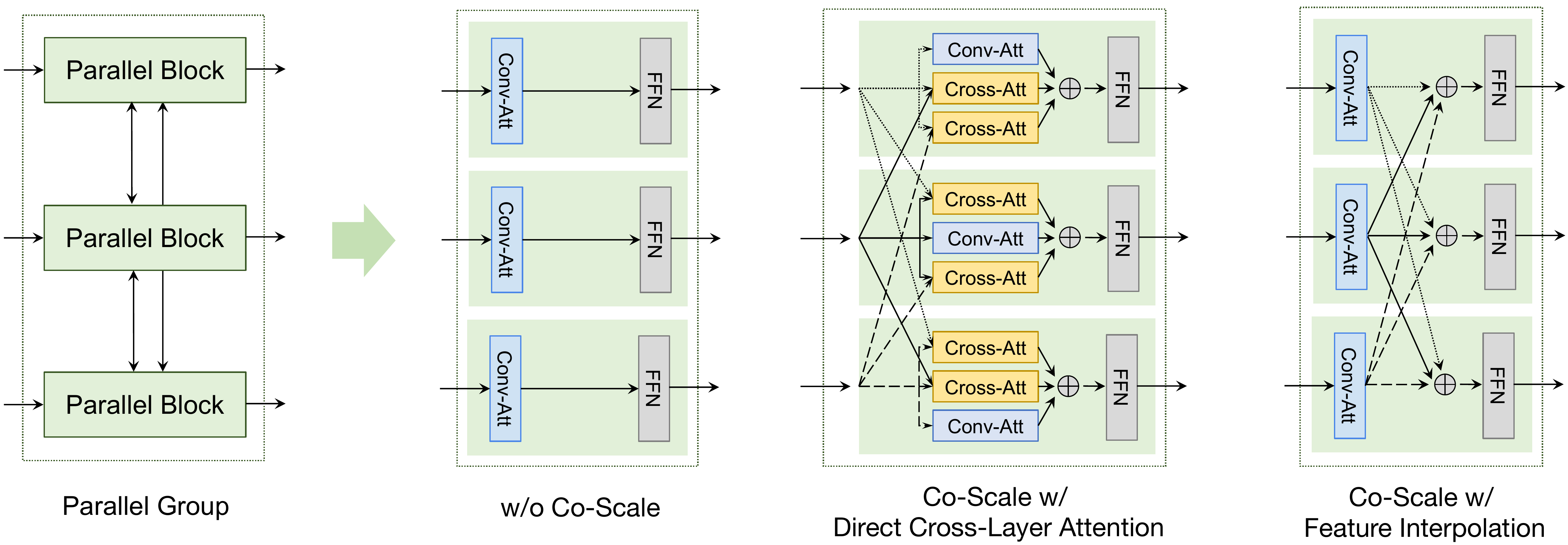}
\caption{\textbf{Schematic illustration of the parallel group in CoaT.} For ``w/o Co-Scale'', tokens learned at the individual scales are combined to perform the classification but absent intermediate co-scale interaction for the individual paths of the parallel blocks. We propose two co-scale variants, namely direct cross-layer attention and attention with feature interpolation. Co-scale with feature interpolation is adopted in the final CoaT-Lite and CoaT models reported on the ImageNet benchmark.}
\label{fig:parallel-group}
\end{figure*} 

\begin{table*}[htb]
\centering
\caption{
        \textbf{Architecture details of CoaT-Lite and CoaT models.} $C_{i}$ represents the hidden dimension of the attention layers in block $i$; $H_{i}$ represents the number of attention heads in the attention layers in block $i$; $R_{i}$ represents the expansion ratio for the feed-forward hidden layer dimension between attention layers in block $i$. Multipliers indicate the number of conv-attentional modules in block $i$.
       }
    \label{tab:architecture}
\scalebox{0.62}{
\begin{tabular}{c|c|c|c|c|c|c| c|c}
	\bottomrule[0.15em]
	 \multirow{2}{*}{Blocks} & \multirow{2}{*}{Output} &\multicolumn{4}{c|}{CoaT-Lite} &\multicolumn{3}{c}{CoaT} \\
	 	 \cline{3-9} 
	 & &  Tiny & Mini & Small & Medium & Tiny & Mini  & Small\\ 
   	\hline 
   	& & & & & &  & & \\ [-1.5ex]
    \makecell{Serial Block \\ ($\text{S}_1$)} & $56 \times 56$ & $\begin{bmatrix}
	\begin{array}{l}
	C_1=64 \\
	H_1=8 \\
	R_1=8 \\
	\end{array}
	\end{bmatrix} \times 2$ & 
	$\begin{bmatrix}
	\begin{array}{l}
	C_1=64 \\
	H_1=8 \\
	R_1=8 \\
	\end{array}
	\end{bmatrix} \times 2$ &
	$\begin{bmatrix}
	\begin{array}{l}
	C_1=64\\
	H_1=8 \\
	R_1=8 \\
	\end{array}
	\end{bmatrix} \times 3$ &
	$\begin{bmatrix}
	\begin{array}{l}
	C_1=128\\
	H_1=8 \\
	R_1=4 \\
	\end{array}
	\end{bmatrix} \times 3$ &
	$\begin{bmatrix}
	\begin{array}{l}
	C_1=152\\
	H_1=8 \\
	R_1=4 \\
	\end{array}
	\end{bmatrix} \times 2$ &
	$\begin{bmatrix}
	\begin{array}{l}
	C_1=152\\
	H_1=8 \\
	R_1=4 \\
	\end{array}
	\end{bmatrix} \times 2$ &
	$\begin{bmatrix}
	\begin{array}{l}
	C_1=152\\
	H_1=8 \\
	R_1=4 \\
	\end{array}
	\end{bmatrix} \times 2$
	\\
	& & & & & & & &\\ [-1.5ex]
	\hline
	& & & & & & & &\\ [-1.5ex]
	\makecell{Serial Block \\ ($\text{S}_2$)} & $28 \times 28$ & $\begin{bmatrix}
	\begin{array}{l}
	C_2=128 \\
	H_2=8 \\
	R_2=8 \\
	\end{array}
	\end{bmatrix} \times 2$ &
	$\begin{bmatrix}
	\begin{array}{l}
	C_2=128 \\
	H_2=8 \\
	R_2=8 \\
	\end{array}
	\end{bmatrix} \times 2$ &
	$\begin{bmatrix}
	\begin{array}{l}
	C_2=128 \\
	H_2=8 \\
	R_2=8 \\
	\end{array}
	\end{bmatrix} \times 4$ &
	$\begin{bmatrix}
	\begin{array}{l}
	C_1=256\\
	H_1=8 \\
	R_1=4 \\
	\end{array}
	\end{bmatrix} \times 6$ &
	$\begin{bmatrix}
	\begin{array}{l}
	C_2=152 \\
	H_2=8 \\
	R_2=4 \\
	\end{array}
	\end{bmatrix} \times 2$ &
	$\begin{bmatrix}
	\begin{array}{l}
	C_2=216 \\
	H_2=8 \\
	R_2=4 \\
	\end{array}
	\end{bmatrix} \times 2$ &
	$\begin{bmatrix}
	\begin{array}{l}
	C_1=320\\
	H_1=8 \\
	R_1=4 \\
	\end{array}
	\end{bmatrix} \times 2$ 
	\\
	& & & & & & & &\\ [-1.5ex]
	\hline
	& & & & & & & &\\ [-1.5ex]
	\makecell{Serial Block \\ ($\text{S}_3$)} & $14 \times 14$ & $\begin{bmatrix}
	\begin{array}{l}
	C_3=256 \\
	H_3=8 \\
	R_3=4 \\
	\end{array}
	\end{bmatrix} \times 2$ &
	$\begin{bmatrix}
	\begin{array}{l}
	C_3=320 \\
	H_3=8 \\
	R_3=4 \\
	\end{array}
	\end{bmatrix} \times 2$ &
	$\begin{bmatrix}
	\begin{array}{l}
	C_3=320 \\
	H_3=8 \\
	R_3=4 \\
	\end{array}
	\end{bmatrix} \times 6$ &
	$\begin{bmatrix}
	\begin{array}{l}
	C_1=320\\
	H_1=8 \\
	R_1=4 \\
	\end{array}
	\end{bmatrix} \times 10$ &
	$\begin{bmatrix}
	\begin{array}{l}
	C_3=152 \\
	H_3=8 \\
	R_3=4 \\
	\end{array}
	\end{bmatrix} \times 2$ &
	$\begin{bmatrix}
	\begin{array}{l}
	C_3=216 \\
	H_3=8 \\
	R_3=4 \\
	\end{array}
	\end{bmatrix} \times 2$ &
	$\begin{bmatrix}
	\begin{array}{l}
	C_1=320\\
	H_1=8 \\
	R_1=4 \\
	\end{array}
	\end{bmatrix} \times 2$ \\
	& & & & & & & &\\ [-1.5ex]
	\hline
    & & & & & & & & \\ [-1.5ex]
	\makecell{Serial Block \\ ($\text{S}_4$)} & $7 \times 7$ &$\begin{bmatrix}
	\begin{array}{l}
	C_4=320 \\
	H_4=8 \\
	R_4=4 \\
	\end{array}
	\end{bmatrix} \times 2$& 
	$\begin{bmatrix}
	\begin{array}{l}
	C_4=512 \\
	H_4=8 \\
	R_4=4 \\
	\end{array}
	\end{bmatrix} \times 2$ &
	$\begin{bmatrix}
	\begin{array}{l}
	C_4=512 \\
	H_4=8 \\
	R_4=4 \\
	\end{array}
	\end{bmatrix} \times 3$ &
	$\begin{bmatrix}
	\begin{array}{l}
	C_1=512\\
	H_1=8 \\
	R_1=4 \\
	\end{array}
	\end{bmatrix} \times 8$ &
	$\begin{bmatrix}
	\begin{array}{l}
	C_4=152 \\
	H_4=8 \\
	R_4=4 \\
	\end{array}
	\end{bmatrix} \times 2$ & 
	$\begin{bmatrix}
	\begin{array}{l}
	C_4=216 \\
	H_4=8 \\
	R_4=4 \\
	\end{array}
	\end{bmatrix} \times 2$&
	$\begin{bmatrix}
	\begin{array}{l}
	C_1=320\\
	H_1=8 \\
	R_1=4 \\
	\end{array}
	\end{bmatrix} \times 2$\\
	& & & & & & & &\\ [-1.5ex]
	\hline
    & & & & & & & & \\ [-1.5ex]
	\makecell{Parallel Group} &
	$\begin{bmatrix}
	\begin{array}{l}
	 28 \times 28   \\
	 14 \times 14 \\
	 \,\,\, 7 \times 7 \\
	\end{array}
	\end{bmatrix}$
	 & &  &  & &
	$\begin{bmatrix}
	\begin{array}{l}
	C_4=152 \\
	H_4=8 \\
	R_4=4 \\
	\end{array}
	\end{bmatrix} \times 6$ & 
	$\begin{bmatrix}
	\begin{array}{l}
	C_4=216 \\
	H_4=8 \\
	R_4=4 \\
	\end{array}
	\end{bmatrix} \times 6$  &
	$\begin{bmatrix}
	\begin{array}{l}
	C_1=320\\
	H_1=8 \\
	R_1=4 \\
	\end{array}
	\end{bmatrix} \times 6$ \\ [+3.8ex] 

	\hline
	\multicolumn{2}{c|}{} & & & & & \\ [-2.5ex]
	\multicolumn{2}{c|}{\#Params} &5.7M& 11M&20M &45M &5.5M &10M &22M \\ [-0.4ex]
\bottomrule[0.15em]
\end{tabular}
 }
\vspace{2mm}
\end{table*}

\vspace{-3mm}
\paragraph{Convolutional Position Encoding.} We then extend the idea of convolutional relative position encoding to a general convolutional position encoding case. Convolutional relative position encoding models local position-based relationships between queries and values. Similar to the absolute position encoding used in most image Transformers \cite{dosovitskiy2020image,touvron2020training}, we would like to insert the position relationship into the input image features directly to enrich the effects of relative position encoding. In each conv-attentional module, we insert a depthwise convolution into the input features $X$ and concatenate the resulting position-aware features back to the input features following the standard absolute position encoding scheme (see Figure~\ref{fig:conv-att} lower part), which resembles the realization of conditional position encoding in CPVT \cite{chu2021we}.

CoaT and CoaT-Lite share the convolutional position encoding weights and convolutional relative position encoding weights for the serial and parallel modules within the same scale. We set convolution kernel size to 3 for the convolutional position encoding. We set convolution kernel size to 3, 5 and 7 for image features from different attention heads for convolutional relative position encoding. 

The work of CPVT \cite{chu2021we} explores the use of convolution as conditional position encodings by inserting it after the feed-forward network under a single scale ($\frac{H}{16}\times\frac{W}{16}$). Our work focuses on applying convolution as relative position encoding and a general position encoding with the factorized attention.
\vspace{-3mm}
\paragraph{Conv-Attentional Mechanism}

The final conv-attentional module is shown in Figure \ref{fig:conv-att}: We apply the first convolutional position encoding on the image tokens from the input. Then, we feed it into $\text{ConvAtt}(\cdot)$ including factorized attention and the convolutional relative position encoding. The resulting map is used for the subsequent feed-forward networks.

\section{Co-Scale Conv-Attentional Transformers}

\subsection{Co-Scale Mechanism}
The proposed co-scale mechanism is designed to introduce fine-to-coarse, coarse-to-fine and cross-scale information into image transformers. Here, we describe two types of building blocks in the CoaT architecture, namely serial and parallel blocks, in order to model multiple scales and enable the co-scale mechanism.
\vspace{-3mm}
\paragraph{CoaT Serial Block.}
A serial block (shown in Figure \ref{fig:serial-block}) models image representations in a reduced resolution. In a typical serial block, we first down-sample input feature maps by a certain ratio using a patch embedding layer, and flatten the reduced feature maps into a sequence of image tokens. We then concatenate image tokens with an additional \texttt{CLS} token, a specialized vector to perform image classification, and apply multiple conv-attentional modules as described in Section \ref{sec:conv_att} to learn internal relationships among image tokens and the \texttt{CLS} token. Finally, we separate the \texttt{CLS} token from the image tokens and reshape the image tokens to 2-D feature maps for the next serial block. 
\vspace{-3mm}
\paragraph{CoaT Parallel Block.}
We realize a co-scale mechanism between parallel blocks in each parallel group (shown in Figure \ref{fig:parallel-group}). In a typical parallel group, we have sequences of input features (image tokens and \texttt{CLS} token) from serial blocks with different scales. To enable fine-to-coarse, coarse-to-fine, and cross-scale interaction in the parallel group, we develop two strategies: (1) direct cross-layer attention; (2) attention with feature interpolation. In this paper, we adopt attention with feature interpolation for better empirical performance. The effectiveness of both strategies is shown in Section \ref{sec:ablation}.

\noindent {\em Direct cross-layer attention.}
In direct cross-layer attention, we form query, key, and value vectors from input features for each scale. For attention within the same layer, we use the conv-attention (Figure \ref{fig:conv-att}) with the query, key and value vectors from current scale. For attention across different layers, we down-sample or up-sample the key and value vectors to match the resolution of other scales, which enables fine-to-coarse and coarse-to-fine interaction. We then perform cross-attention, which extends the conv-attention with queries from the current scale with keys and values from another scale. Finally, we sum the outputs of conv-attention and cross-attention together and apply a shared feed-forward layer. With direct cross-layer attention, the cross-scale information is fused in a cross-attention fashion.

\noindent {\em Attention with feature interpolation.}
Instead of performing cross-layer attention directly, we also present attention with feature interpolation. First, the input image features from different scales are processed by independent conv-attention modules. Then, we down-sample or up-sample image features from each scale to match the dimensions of other scales using bilinear interpolation, or keep the same for its own scale. The features belonging to the same scale are summed in the parallel group, and they are further passed into a shared feed-forward layer. In this way, the conv-attentional module in the next step can learn cross-scale information based on the feature interpolation in the current step.

\subsection{Model Architecture}

\paragraph{CoaT-Lite.}
CoaT-Lite, Figure \ref{fig:pipeline} (Left), processes input images with a series of serial blocks following a fine-to-coarse pyramid structure. Given an input image  $I \in \R^{H \times W \times C}$, each serial block down-samples the image features into lower resolution, resulting in a sequence of four resolutions:$F_1 \in \R^{\frac{H}{4} \times \frac{W}{4} \times C_{1}}$, $F_2 \in \R^{\frac{H}{8} \times \frac{W}{8} \times C_{2}}$, $F_3 \in \R^{\frac{H}{16} \times \frac{W}{16} \times C_{3}}$, $F_4 \in \R^{\frac{H}{32} \times \frac{W}{32} \times C_{4}}$. In CoaT-Lite, we obtain the \texttt{CLS} token in the last serial block, and perform image classification via a linear projection layer based on the \texttt{CLS} token.
\vspace{-2mm}
\paragraph{CoaT.}
Our CoaT model, shown in Figure \ref{fig:pipeline} (Right), consists of both serial and parallel blocks. Once we obtain multi-scale feature maps $\{F_1, F_2, F_3, F_4\}$ from the serial blocks, we pass $F_2, F_3, F_4$ and corresponding \texttt{CLS} tokens into the parallel group with three separate parallel blocks. To perform classification with CoaT, we aggregate the \texttt{CLS} tokens from all three scales.
\vspace{-3mm}
\paragraph{Model Variants.}
In this paper, we explore CoaT and CoaT-Lite with several different model sizes, namely Tiny, Mini, Small and Medium. Architecture details are shown in Table \ref{tab:architecture}. For example, tiny models represent those with a 5M parameter budget constraint. Specifically, these tiny models have four serial blocks, each with two conv-attentional modules. In CoaT-Lite Tiny architectures, the hidden dimensions of the attention layers increase in later blocks. CoaT Tiny sets the hidden dimensions of the attention layers in the parallel group to be equal, and performs the co-scale mechanism within the six parallel groups. Mini, small and medium models follow the same architecture design but with increased embedding dimensions and increased numbers of conv-attentional modules within blocks.

\begin{table}
    \centering
    \caption{
    \textbf{CoaT performance on ImageNet-1K validation set.} Our CoaT models consistently outperform other methods while being parameter efficient. ConvNets and ViTNets with similar model size are grouped together for comparison.  ``\#GFLOPs" and Top-1 Acc are measured at input image size. ``*" results are adopted from \cite{wang2021pyramid}.
    }
    \label{tab:imagenet}
    \vspace{1mm}
    \resizebox{0.48\textwidth}{!}{
        \begin{tabular}{c | l|c|cc|c}
        \toprule [0.15em]
        \bf Arch. & \bf  Model  & \bf  \#Params &  \bf  Input  & \bf \#GFLOPs &  \bf Top-1 Acc.  \\
        \midrule [0.1em]
        ConvNets  & EfficientNet-B0 \cite{tan2019efficientnet} & 5.3M  & $224^{2}$    & 0.4     & 77.1\%        \\
                  & ShuffleNet  \cite{zhang2018shufflenet}     & 5.4M  & $224^{2}$    & 0.5     & 73.7\%        \\
        \midrule
        ViTNets   & DeiT-Tiny  \cite{touvron2020training}      & 5.7M    & $224^{2}$  & 1.3     & 72.2\%        \\
                  & CPVT-Tiny \cite{chu2021we}                 & 5.7M    & $224^{2}$  & -       & 73.4\%        \\
   \rowcolor{mygray}
   & \textbf{CoaT-Lite} \footnotesize{\textbf{Tiny} (Ours)}    & 5.7M    & $224^{2}$  & 1.6     & 77.5\%        \\
   \rowcolor{mygray}
   & \textbf{CoaT}   \footnotesize{\textbf{Tiny} (Ours)}       & 5.5M    & $224^{2}$  & 4.4     & \bf 78.3\%    \\
        \midrule [0.1em]
        ConvNets  & EfficientNet-B2\cite{tan2019efficientnet}  & 9M      & $260^{2}$  & 1.0     & 80.1\%        \\    
                  & ResNet-18$* $  \cite{he2016deep}           & 12M     & $224^{2}$  & 1.8     & 69.8\%        \\
        \midrule
        ViTNets   & PVT-Tiny  \cite{wang2021pyramid}           & 13M     & $224^{2}$  & 1.9     & 75.1\%        \\
   \rowcolor{mygray}
   & \textbf{CoaT-Lite}   \footnotesize{\textbf{Mini} (Ours)}  & 11M     & $224^{2}$  & 2.0     & 79.1\%        \\
   \rowcolor{mygray}
   & \textbf{CoaT}      \footnotesize{\textbf{Mini} (Ours)}    & 10M     & $224^{2}$  & 6.8     & \bf 81.0\%    \\
        \midrule [0.1em]
		ConvNets  & EfficientNet-B4 \cite{tan2019efficientnet} & 19M     & $380^{2}$  & 4.2     & \bf 82.9\%    \\
                  & ResNet-50$* $   \cite{he2016deep}          & 25M     & $224^{2}$  & 4.1     & 78.5\%        \\
                  & ResNeXt50-32x4d* \cite{xie2017aggregated}  & 25M     & $224^{2}$  & 4.3     & 79.5\%        \\
		\midrule
		ViTNets   & DeiT-Small  \cite{touvron2020training}     & 22M     & $224^{2}$  & 4.6     & 79.8\%        \\
                  & PVT-Small  \cite{wang2021pyramid}          & 24M     & $224^{2}$  & 3.8     & 79.8\%        \\
                  & CPVT-Small   \cite{chu2021we}              & 22M     & $224^{2}$  & -       & 80.5\%        \\
                  & T2T-ViT$_{t}$-14   \cite{yuan2021tokens}   & 22M     & $224^{2}$  & 6.1     & 81.7\%        \\ 
                  & Swin-T  \cite{liu2021swin}                 & 29M     & $224^{2}$  & 4.5     & 81.3\%        \\ 
                    
   \rowcolor{mygray}
   & \textbf{CoaT-Lite}  \footnotesize{\textbf{Small} (Ours)}  & 20M     & $224^{2}$  & 4.0     & 81.9\%        \\
   \rowcolor{mygray}
   & \textbf{CoaT}  \footnotesize{\textbf{Small} (Ours)}       & 22M     & $224^{2}$  & 12.6    & 82.1\%        \\
        \midrule [0.1em]
        ConvNets  & EfficientNet-B6 \cite{tan2019efficientnet} & 43M     & $528^{2}$  & 19      & 84.0\%        \\
		          & ResNet-101$* $ \cite{he2016deep}           & 45M     & $224^{2}$  & 7.9     & 79.8\%        \\
		          & ResNeXt101-64x4d$* $ \cite{xie2017aggregated} & 84M  & $224^{2}$  & 15.6    & 81.5\%        \\
		\midrule 
        ViTNets & PVT-Large  \cite{wang2021pyramid}            & 61M     & $224^{2}$  & 9.8     & 81.7\%        \\
		            & T2T-ViT$_{t}$-24   \cite{yuan2021tokens} & 64M     & $224^{2}$  & 15      & 82.6\%        \\
		            & DeiT-Base     \cite{touvron2020training} & 86M     & $224^{2}$  & 17.6    & 81.8\%        \\
                    & CPVT-Base    \cite{chu2021we}            & 86M     & $224^{2}$  & -       & 82.3\%        \\
                    & Swin-B \cite{liu2021swin}                & 88M     & $224^{2}$  & 15.4    & 83.5\%        \\ 
                    & Swin-B \cite{liu2021swin}                & 88M     & $384^{2}$  & 47      & \bf 84.5\%    \\ 
                    
   \rowcolor{mygray}
   & \textbf{CoaT-Lite}  \footnotesize{\textbf{Medium} (Ours)} & 45M     & $224^{2}$  & 9.8     &  83.6\%        \\
   \rowcolor{mygray}
   & \textbf{CoaT-Lite}  \footnotesize{\textbf{Medium} (Ours)} & 45M     & $384^{2}$  & 28.7    & \bf 84.5\%     \\
   
        \bottomrule[0.15em]
        \end{tabular}
  }  
  \vspace{-4mm}
\end{table}

\begin{table}[!htb]
\centering
\caption{\textbf{Object detection and instance segmentation results based on Mask R-CNN on COCO val2017}. Experiments are performed under the MMDetection framework \cite{mmdetection}. ``*'' results are adopted from Detectron2.}
\label{tab:maskrcnn}
\resizebox{0.40\textwidth}{!}{
\begin{tabular}{l|c|cc|cc}
\multirow{2}{*}{Backbone} & \multirow{2}{*}{\tabincell{c}{\#Params \\(M)}} &\multicolumn{2}{c|}{ w/ FPN 1$\times$} &\multicolumn{2}{c}{ w/ FPN 3$\times$} \\
\cline{3-6} 
& &AP$^{\rm b}$ &AP$^{\rm m}$ &AP$^{\rm b}$ & AP$^{\rm m}$  \\
\bottomrule[0.10em]
\hline
ResNet-18*                                              & 31.3 & 34.2 & 31.3 & 36.3 & 33.2 \\
PVT-Tiny \cite{wang2021pyramid}                         & 32.9 & 36.7 & 35.1 & 39.8 & 37.4 \\
\rowcolor{mygray}
\textbf{CoaT-Lite} \footnotesize{\textbf{Mini} (Ours)}  & 30.7 & 41.4 & 38.0 & 42.9 & 38.9 \\
\rowcolor{mygray}
\textbf{CoaT} \footnotesize{\textbf{Mini} (Ours)}       & 30.2 & \textbf{45.1} & \textbf{40.6} & \textbf{46.5} & \textbf{41.8} \\
\hline
ResNet-50*                                              & 44.3 & 38.6 & 35.2 & 41.0 & 37.2 \\
PVT-Small \cite{wang2021pyramid}                        & 44.1 & 40.4 & 37.8 & 43.0 & 39.9 \\
Swin-T \cite{liu2021swin}                               & 47.8 & 43.7 & 39.8 & 46.0 & 41.6  \\
\rowcolor{mygray}
\textbf{CoaT-Lite} \footnotesize{\textbf{Small} (Ours)} & 39.5 & 45.2 & 40.7 & 45.7 & 41.1  \\
\rowcolor{mygray}
\textbf{CoaT} \footnotesize{\textbf{Small} (Ours)}      & 41.6 & \textbf{46.5} & \textbf{41.8} &\textbf{49.0} & \textbf{43.7} \\
\end{tabular}
}
\vspace{-2mm}
\end{table}

\begin{table}[!htb]
\centering
\caption{\textbf{Object detection and instance segmentation results based on Cascade Mask R-CNN  on COCO val2017}. Experiments are performed using the MMDetection framework \cite{mmdetection}.}
\vspace{3mm}
\label{tab:cascade-maskrcnn}
\resizebox{0.40\textwidth}{!}{
\begin{tabular}{l|c|cc|cc}
\multirow{2}{*}{Backbone} & \multirow{2}{*}{\tabincell{c}{\#Params \\(M)}} &\multicolumn{2}{c|}{ w/ FPN 1$\times$} &\multicolumn{2}{c}{ w/ FPN 3$\times$} \\
\cline{3-6} 
& &AP$^{\rm b}$ &AP$^{\rm m}$ &AP$^{\rm b}$ & AP$^{\rm m}$  \\
\bottomrule[0.10em]
\hline
Swin-T \cite{liu2021swin}                                 & 85.6 & 48.1 & 41.7 & 50.4 & 43.7 \\
\rowcolor{mygray}
\textbf{CoaT-Lite} \footnotesize{\textbf{Small} (Ours)}   & 77.3 & 49.1 & 42.5 & 48.9 & 42.6 \\
\rowcolor{mygray}
\textbf{CoaT} \footnotesize{\textbf{Small} (Ours)}        & 79.4 & \textbf{50.4} & \textbf{43.5} & \textbf{52.2} & \textbf{45.1} \\
\end{tabular}
}
\vspace{-2mm}
\end{table}

\begin{table}[!htp]
\centering
\caption{\textbf{Object detection results based on Deformable DETR on COCO val2017}. DD ResNet-50 represents the baseline result using the official checkpoint. ResNet-50 and our CoaT-Lite as DD backbones are directly comparable due to similar model size.}
\vspace{2mm}
\label{tab:deformable-detr}
\resizebox{0.45\textwidth}{!}{
\begin{tabular}{l|lcc|ccc}
	\multirow{2}{*}{Backbone} & \multicolumn{6}{c}{Deformable DETR (Multi-Scale)}\\
	\cline{2-7}
    &AP &AP$_{50}$ &AP$_{75}$ &AP$_S$ &AP$_M$ &AP$_L$  \\
    \bottomrule[0.10em]
	DD ResNet-50 \cite{zhu2020deformable}                        & 44.5 & 63.7 & 48.7 & 26.8 & 47.6 & 59.6 \\

	\rowcolor{mygray}
	DD \textbf{CoaT-Lite}  \footnotesize{\textbf{Small} (Ours)}  & 47.0 & 66.5 & 51.2 & 28.8 & 50.3 & 63.3 \\
    DD \textbf{CoaT}  \footnotesize{\textbf{Small} (Ours)}       & \textbf{48.4} & \textbf{68.5} & \textbf{52.4} & \textbf{30.2} & \textbf{51.8} & \textbf{63.8} \\
\end{tabular}
}
\vspace{-6mm}
\end{table}

\section{Experiments}

\subsection{Experiment Details}

\paragraph{Image Classification.}
We perform image classification on the standard ILSVRC-2012 ImageNet dataset \cite{russakovsky2015imagenet}.
The standard ImageNet benchmark contains 1.3 million images in the training set and 50K images in the validation set, covering 1000 object classes.
Image cropping sizes are set to 224$\times$224. For fair comparison, we perform data augmentation such as MixUp \cite{zhang2017mixup}, CutMix \cite{yun2019cutmix}, random erasing \cite{zhong2020random}, repeated augmentation \cite{hoffer2020augment}, and label smoothing \cite{szegedy2016rethinking}, following identical procedures in DeiT \cite{touvron2020training}.

All experimental results for our models in Table \ref{tab:imagenet} are reported at 300 epochs, consistent with previous methods \cite{touvron2020training}. All models are trained with the AdamW \cite{loshchilov2017decoupled} optimizer under the NVIDIA Automatic Mixed Precision (AMP) framework. The learning rate is scaled as $5\times10^{-4}\times \frac{\text{global batch size}}{512}$.

\vspace{-5mm} 

\paragraph{Object Detection and Instance Segmentation.} 
We conduct object detection and instance segmentation experiments on the Common Objects in Context (COCO2017) dataset \cite{lin2014microsoft}. The COCO2017 benchmark contains 118K training images and 5K validation images. We evaluate the generalization ability of CoaT in object detection and instance segmentation with the Mask R-CNN \cite{he2017mask} and Cascade Mask R-CNN \cite{cai2019cascade}. We use the MMDetection \cite{mmdetection} framework and follow the settings from Swin Transformers \cite{liu2021swin}. In addition, we perform object detection based on Deformable DETR \cite{zhu2020deformable} following its data processing settings.

For Mask R-CNN optimization, we train the model with the ImageNet-pretrained backbone on 8 GPUs via AdamW optimizer with learning rate 0.0001. The training period contains 12 epochs in 1$\times$ setting and 36 epochs in 3$\times$ setting. For Cascade R-CNN experiments, we use three detection heads, with the same optimization and training period as Mask R-CNN. For Deformable DETR optimization, we train the model with the pretrained backbone for 50 epochs, using an AdamW optimizer with initial learning rate $2\times10^{-4}$, $\beta_{1}=0.9$, and $\beta_{2}=0.999$. We reduce the learning rate by a factor of 10 at epoch 40.

\subsection{CoaT for ImageNet Classification}

Table \ref{tab:imagenet} shows top-1 accuracy results for our models on the ImageNet validation set comparing with state-of-the-art methods. We separate model architectures into two categories: convolutional networks (ConvNets), and Transformers (ViTNets). Under different parameter budget constraints, CoaT and CoaT-Lite show strong results compared to other ConvNet and ViTNet methods.

\subsection{Object Detection and Instance Segmentation}

Tables \ref{tab:maskrcnn} and \ref{tab:cascade-maskrcnn} demonstrate CoaT object detection and instance segmentation results under the Mask R-CNN and Cascade Mask R-CNN frameworks on the COCO val2017 dataset. Our CoaT and CoaT-Lite models show clear performance advantages over the ResNet, PVT \cite{wang2021pyramid} and Swin \cite{liu2021swin} backbones under both the 1$\times$ setting and the 3$\times$ setting. In particular, our CoaT models bring a large performance gain, demonstrating that our co-scale mechanism is essential to improve the performance of Transformer-based architectures for downstream tasks.  

We additionally perform object detection with the Deformable DETR (DD) framework in Table \ref{tab:deformable-detr}. We compare our models with the standard ResNet-50 backbone on the COCO dataset \cite{lin2014microsoft}. Our CoaT backbone achieves 3.9\% improvement on average precision (AP) over the results of Deformable DETR with ResNet-50 \cite{zhu2020deformable}.

\subsection{Ablation Study}
\label{sec:ablation}

\paragraph{Effectiveness of Position Encodings.} We study the effectiveness of the combination of the convolutional relative position encoding (CRPE) and convolutional position encoding (CPE) in our conv-attention module in Table \ref{tab:cpe-crpe}. Our CoaT-Lite without any position encoding results in poor performance, indicating that position encoding is essential for vision Transformers. We observe great improvement for CoaT-Lite variants with either CRPE or CPE, and the combination of CRPE and CPE leads to the best performance (77.5\% top-1 accuracy), making both position encoding schemes complementary rather than conflicting. \\
\begin{table}[!htp]
    \centering
    \caption{
    \textbf{Effectiveness of position encodings.} All experiments are performed with the CoaT-Lite Tiny architecture. Performance is evaluated on the ImageNet-1K validation set. }
    \label{tab:cpe-crpe}
    \vspace{1mm}
    \resizebox{0.30\textwidth}{!}{
        \begin{tabular}{l|c|c|c}
        \toprule [0.15em]
         Model         & CPE                    &  CRPE              &  Top-1 Acc.   \\
        \midrule [0.1em]
        CoaT-Lite Tiny &   \xmark               &  \xmark            & 68.8\%      \\
                       &   \xmark               &  \cmark            & 75.0\%      \\
                       &   \cmark               &  \xmark            & 75.9\%      \\
                       &   \cmark               &  \cmark            & 77.5\%      \\
        \bottomrule[0.15em]
        \end{tabular}
    }
    \vspace{-4mm}
\end{table}

\vspace{-5mm}

\paragraph{Effectiveness of Co-Scale.} In Table \ref{tab:co-scale}, we present performance results for two co-scale variants in CoaT, direct cross-layer attention and attention with feature interpolation. We also report CoaT without co-scale as a baseline. Comparing to CoaT without a co-scale mechanism, CoaT with feature interpolation shows performance improvements on both image classification and object detection (Mask R-CNN w/ FPN 1$\times$). Attention with feature interpolation offers a clear advantage over direct cross-layer attention due to less computational complexity and higher accuracy.    

\begin{table}[!htp]
    \centering
    \caption{
    \textbf{Effectiveness of co-scale.} All experiments are performed with the CoaT Tiny architecture. Performance is evaluated on the ImageNet-1K validation set and the COCO val2017 dataset. }
    \label{tab:co-scale}
    \vspace{1mm}
    \resizebox{0.48\textwidth}{!}{
        \begin{tabular}{ l|c cc|ccc}
        \toprule [0.15em]
         Model  & \#Params &  Input  &\#GFLOPs &  Top-1 Acc. @input  &AP$^{\rm b}$ &AP$^{\rm m}$ \\
        \midrule [0.1em]
        CoaT w/o Co-Scale                           & 5.5M  & $224^{2}$  & 4.4     & 77.8\%    & 41.6 & 37.9 \\
        CoaT w/ Co-Scale                            & & & &      \\
            \quad - Direct Cross-Layer Attention    & 5.5M  & $224^{2}$  & 4.8     & 77.0\%    & 42.1 & 38.3 \\
            \quad - Attention w/ Feature Interp.    & 5.5M  & $224^{2}$  & 4.4     & 78.3\%    & 42.5 & 38.6 \\
        \bottomrule[0.15em]

        \end{tabular}
    }
    \vspace{-8mm}
\end{table}

\paragraph{Computational Cost.} We report FLOPs, FPS, latency, and GPU memory usage in Table \ref{tab:computation}. In summary, CoaT models attain higher accuracy than similar-sized Swin Transformers, but CoaT models in general do have larger latency/FLOPs. The current parallel groups in CoaT are more computationally demanding, which can be mitigated by reducing high-resolution parallel blocks and re-using their feature maps in the co-scale mechanism in future work. The latency overhead in CoaT is possibly because operations (e.g. layers, position encodings, upsamples/downsamples) are not running in parallel.
\begin{table}[h]
    \centering
    \caption{\scriptsize{\textbf{ImageNet-1K validation set results} compared with the concurrent work Swin Transformer\cite{liu2021swin}. Computational metrics are measured on a single V100 GPU.}}
    \label{tab:computation}
    \resizebox{0.48\textwidth}{!}{
        \vspace{-2mm}
        \begin{tabular}{l|c|ccccc|cc}
        \toprule [0.15em]
        Model  & \#Params &  Input  & GFLOPs & FPS & Latency & Mem &  Top-1 Acc.  &  Top-5 Acc. \\
        \midrule [0.1em]
        Swin-T \cite{liu2021swin}                                  & 28M  & $224^{2}$   & 4.5   & 755 & 16ms & 222M & 81.2\% & 95.5\%  \\
        \rowcolor{mygray}
        \textbf{CoaT-Lite}  \footnotesize{\textbf{Small} (Ours)}   & 20M  & $224^{2}$   & 4.0   & 634 & 32ms & 224M & 81.9\% & 95.6\% \\
        \rowcolor{mygray}
        \textbf{CoaT}  \footnotesize{\textbf{Small} (Ours)}        & 22M  & $224^{2}$   & 12.6  & 111 & 60ms & 371M & \bf 82.1\% & \bf 96.1\% \\
        \midrule [0.1em]

        Swin-S \cite{liu2021swin}                                  & 50M  & $224^{2}$   & 8.7   & 437 & 29ms & 372M & 83.2\% & 96.2\% \\
        Swin-B \cite{liu2021swin}                                  & 88M  & $224^{2}$   & 15.4  & 278 & 30ms & 579M & 83.5\% & 96.5\% \\        
        \rowcolor{mygray}
        \textbf{CoaT-Lite}  \footnotesize{\textbf{Medium} (Ours)}  & 45M  & $224^{2}$   & 9.8   & 319 & 52ms & 429M & \bf 83.6\% & \bf 96.7\% \\
        \midrule [0.1em]
        
        Swin-B \cite{liu2021swin}                                  & 88M  & $384^{2}$   & 47.1  & 85  & 33ms & 1250M & \bf 84.5\% & 97.0\% \\ 
        \rowcolor{mygray}
        \textbf{CoaT-Lite}  \footnotesize{\textbf{Medium} (Ours)}  & 45M  & $384^{2}$   & 28.7  & 97  & 56ms & 937M & \bf 84.5\% & \bf 97.1\% \\
        
        \bottomrule[0.15em]
         
        \end{tabular}
    }
        \vspace{-3mm}
\end{table}

\vspace{-2mm}
\section{Conclusion}
\vspace{-2mm}

In this paper, we present a Transformer based image classifier, Co-scale conv-attentional image Transformer (CoaT), in which cross-scale attention and efficient conv-attention operations have been developed. CoaT models attain strong classification results on ImageNet, and their applicability to downstream computer vision tasks has been demonstrated for object detection and instance segmentation.

\paragraph{Acknowledgments.} This work is supported by NSF Award IIS-1717431. Tyler Chang is partially supported by the UCSD HDSI graduate fellowship.

{\small
\bibliographystyle{ieee_fullname}
\bibliography{egbib}
}

\end{document}